\documentclass{article}
\pdfoutput=1 

    \usepackage[preprint]{neurips_2022}

\usepackage{array}

\usepackage[utf8]{inputenc} % allow utf-8 input
\usepackage[T1]{fontenc}    % use 8-bit T1 fonts
\usepackage{hyperref}       % hyperlinks
\usepackage{url}            % simple URL typesetting
\usepackage{booktabs}       % professional-quality tables
\usepackage{amsfonts}       % blackboard math symbols
\usepackage{nicefrac}       % compact symbols for 1/2, etc.
\usepackage{microtype}      % microtypography
\usepackage{xcolor}         % colors
\usepackage{subfig}
\usepackage{multirow}
\usepackage{placeins}
\usepackage{pgfplots}
\pgfplotsset{compat=1.17}

\title{Towards Optimal VPU Compiler Cost Modeling by using Neural Networks to Infer Hardware Performances}

\author{%
  Ian Frederick Vigogne Goodbody Hunter\\
  VPU Presilicon Group\\
  Intel Corporation\\
  Leixlip, Ireland \\
  \texttt{ian.hunter@intel.com} \\
  \And
  Alessandro Palla\\
  VPU Presilicon Group\\
  Intel Corporation\\
  Pisa, Italy\\
  \texttt{alessandro.palla@intel.com} \\
  \And
  Sebastian Eusebiu Nagy\\
  VPU Presilicon Group\\
  Intel Corporation\\
  Timișoara, Romania \\
  \texttt{sebastian.eusebiu.nagy@intel.com} \\
  \And
  Richard Richmond\\
  VPU Presilicon Group\\
  Intel Corporation\\
  Belfast, U.K.\\
  \texttt{richard.richmond@intel.com} \\
  \And
  Kyle McAdoo\\
  VPU Presilicon Group\\
  Intel Corporation\\
  Belfast, U.K.\\
  \texttt{kyle.mcadoo@intel.com} \\
}

\usepackage{graphicx}
\graphicspath{ {./images/} }

\begin{document}

\maketitle

\begin{abstract}

% The abstract must be limited to one paragraph.

Calculating the most efficient schedule of work in a neural network compiler is a difficult task. There are many parameters to be accounted for that can positively or adversely affect that schedule depending on their configuration --- How work is shared between distributed targets, the subdivision of tensors to fit in memory, toggling the enablement of optimizations, etc. Traditionally, neural network compilers determine how to set these values by building a graph of choices and choosing the path with minimal \lq cost \rq. These choices and their corresponding costs are usually determined by an algorithm crafted by engineers with a deep knowledge of the target platform. However, when the amount of options available to a compiler is large, it is very difficult to ensure that these models consistently produce an optimal schedule for all scenarios, whilst still completing compilation in an acceptable timeframe. This paper presents \lq VPUNN\rq --- a neural network-based cost model trained on low-level task profiling that consistently outperforms the state-of-the-art cost modeling in Intel's line of VPU processors. 
\end{abstract}

% Papers may only be up to nine pages long, including figures. % Additional pages containing only acknowledgments and
%references are allowed. Papers that exceed the page limit will not be
%reviewed, or in any other way considered for presentation at the %conference.

% If you wish to post a preprint of your work online, e.g., on arXiv, using the NeurIPS style, please use the \verb+preprint+ option. This will create a nonanonymized version of your work with the text ``Preprint. Work in progress.'' in the footer. This version may be distributed as you see fit. Please \textbf{do not} use the \verb+final+ option, which should \textbf{only} be used for papers accepted to NeurIPS.

% At submission time, please omit the \verb+final+ and \verb+preprint+
%options. This will anonymize your submission and add line numbers to aid review. Please do \emph{not} refer to these line numbers in your paper as they will be removed during generation of camera-ready copies.

\section{Introduction}

\subsection{Overview}

This paper describes a new method of accurately modeling the decision costs within a neural network compiler (henceforth referred to as \lq a compiler\rq ) using a neural network trained on task-level profiling recordings. Specifically, this paper's research describes this new technique as applied to Intel's range of Versatile Processor Units (VPUs) and their corresponding compiler, however, the concepts presented herein can be applied to other compilers, processor types, or other applications that also make use of a cost model (e.g. performance prediction).

The first section of this paper describes existing methods for decision tree resolution and other relevant background information for understanding the context of the research reported in this paper. The following section contains a technical detailing of the VPUNN cost model (and variants thereof). Results gathered during the development are presented and discussed in the penultimate section. Finally, the paper is concluded with a brief discussion on the overall merits of the technique and the possible future direction of this work.

\subsection{Background Information}
\label{background_info}

At its most basic, compiling a neural network for execution on the VPU is a matter of scheduling smaller work items for the processor to complete in an efficient order. This is a common scenario for compilers, independent of research field.

\begin{figure}
  \centering
  \includegraphics[width=0.7\textwidth, bb=-40 0 285 248]{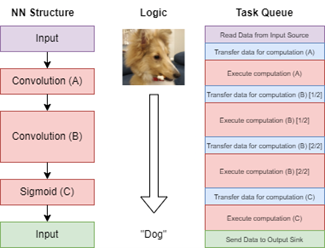}
  % \fbox{\rule[-.5cm]{0cm}{4cm} \rule[-.5cm]{4cm}{0cm}}
  \caption{A sample scheduling of neural network tasks.}
  \label{dog_image}
\end{figure}

\begin{figure}
  \centering
  \includegraphics[width=0.5\textwidth, bb=-20 0 204 63]{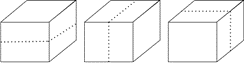}
  \caption{Splitting a tensor on three different axes.}
  \label{split_image}
\end{figure}

Figure \ref{dog_image} shows an example of a hypothetical neural network that classifies an image as either containing a dog or a cat. To execute this network on the VPU, the network is compiled into a queue of low-level tasks such as memory transfers and discrete hardware tensor operations. Scheduling some of these operations can be straightforward --- Convolution A can be completed in a single transfer --- but Convolution B has been split into multiple parts to comply with a limitation of the target platform's memory capacity. This limitation might lead to the compiler subdividing the work items into more compatible sizes. Figure \ref{split_image} shows three options for how this division might be done. In this simple depiction of a 3-dimensional tensor, deciding which axis of division (or \lq split\rq) to choose could be determined by a simple benchmarking of all three options. However, in a real-world scenario where there are more granular division options, optimization toggles, and other significant choices involved, the search space grows rapidly. This is especially true when decisions about multiple adjacent operations are highly correlated with each other. In that case, the inherently combinatorial nature of the problem makes it intractable by brute-forcing all possible decisions.

\begin{figure}
  \centering
  \includegraphics[width=\textwidth, bb=0 0 781 229]{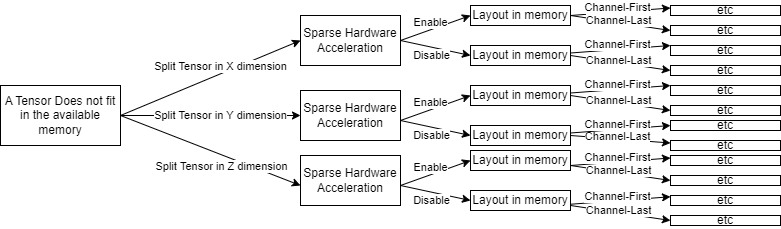}
  \caption{Example of the increasing complexity of a task's decision tree scaling with the amount of options available }
  \label{decision_tree}
\end{figure}

Figure \ref{decision_tree} shows a set of these choices and the resultant decision tree. It can be seen how each additional configurable property markedly increases the number of possible path choices. Because the VPU is an embedded processor with significantly restricted resources (to achieve a low power consumption), the execution schedule can have a very high number of tasks compared to a processor with larger capabilities (e.g. a GPU). This leads to an increased number of compiler decisions to determine how to divide a tensor into transferrable chunks, how much work each distributed unit might consume, configuration of different optimizations (e.g. data compression/decompression, sparse acceleration), etc. all compound to create an exploration space which is unwieldy to resolve in a timely manner.

\subsection{Existing Methods}
\label{existing_methods}

This is not a problem faced solely by VPU compilers and indeed, there is a reasonable body of research dedicated to alleviating the problem. These solutions have primarily taken two forms:

\begin{itemize}
  \item Rule-Based decision: Determining the best scheduling decision by following a series of rules
  \item Cost-based decision: Selecting an outcome based on an algorithm or cost model of the underlying hardware that roughly models the behavior in terms of cycles/power of the target device
\end{itemize}

\subsubsection{Rule Based Methodologies}

Some projects with a decision tree may choose their path based on a series of rules (e.g \citet{Hajek1986KNOWLEDGEBASEDES} describes a rule-based decision system for the best method to maintain a given pavement). An example of a rule might be "Choose to split a tensor on the x-axis if the x-axis is larger than 75 elements". The compiler will test each rule as required until there is only one possible configuration left to choose from. This approach lends itself well to decisions where the choices of schedules are limited and those choices have clear positive and negative effects on the resultant schedule. However, this method becomes difficult to manage when the number of parameters is high, or the effect of each parameter is not significantly pronounced to allow for elimination-style decisions. 

% Hybrid approach to rule/NN methods https://onlinepubs.trb.org/Onlinepubs/trr/1993/1399/1399-001.pdf

\subsubsection{Algorithmic Cost Modeling} 
A more advanced method is for compilers to calculate the final configuration through an algorithmic, or mathematical, methodology. The CoRA Tensor Compiler (as described in \citet{CoRA}) for example operates by computing the schedule which minimizes the amount of memory reloading --- on the grounds that this will minimize wasted computation. A similar modeling function was used as the original cost model for the VPU compiler before VPUNN was integrated. However, this method is only as effective as the correlation of the algorithm to the actual performance of a system.

\subsubsection{Performance Prediction}
Performance prediction is a closely related topic to cost modeling. If the compiler knows the predicted performance of two different decisions, it can pick the preferrable one to proceed with.

\citet{Qi2017PaleoAP} provide algorithmic modeling of a hardware target's performance based upon the arithmetic complexity of the network. Their proposal is a generic approach across multiple hardware targets, though it requires to be provided some high-level information about the target's capabilities and processing paradigms.

More recent work leverages neural networks to model performance. \citet{PredictCost} discuss a prediction system operating on a range of Nvidia GPUs. Similarly in 2018, Intel Habana (\citet{HabanaPaper}) trained a neural network to predict the performance of convolutions and fully-connected layers on the Gaudi processor. However, to the authors' knowledge, the use of these performance predictors as a compiler cost model has not yet been published.

The VPUNN system is differentiated from prior work in several additional aspects

\begin{itemize}
  \item the neural network is trained at a hardware \lq task\rq \  level, rather than at a network \lq operation\rq\  level, allowing for more fine-grained modeling
  \item the VPUNN model includes hardware-specific parameters particular to Intel VPUs
  \item the model supports all available hardware-native operations of the VPU accelerator, rather than a subset.
  \item the VPUNN model is a product-grade cost model for the VPU compiler
\end{itemize}

VPUNN determines the most suitable selection of compiler decisions based on the projected cycle time as computed by its neural network. How this is achieved is described in Section \ref{tech}.

As of the time of writing, the VPUNN system is the cost modeling technology of choice for Intel VPU neural network compilers.

\section{Technical Overview}
\label{tech}

\begin{figure}
  \centering
  \includegraphics[width=0.8\textwidth, bb=-50 0 327 319]{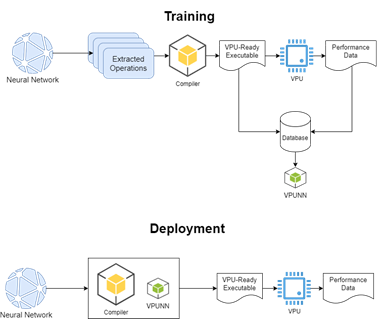}
  \caption{The two phases of VPUNN integration}
  \label{phases_fig}
\end{figure}

A large dataset is key to the successful training of a neural network. Data for training the neural network behind VPUNN was captured using two different methodologies:
\begin{enumerate}
  \item Compiling real-world neural networks into VPU tasks. Each network can generate several hundred tasks, allowing for quick initialization of data points into the database
  \item Creating a wide range of processor tasks through a generation suite. This has been particularly important to unbias the database, as it allows to balance all of the operations and datatype classes.
\end{enumerate}

The generated tasks are then profiled to collect their performance. It is possible to profile individual workloads with very hard granularity using both hardware emulators (Register-transfer level (RTL) simulation, Field-programmable gate array (FPGA) emulation, VPU behavioral model) and directly from the final silicon. This flexibility allows to have a very accurate cost model of the hardware well before silicon tape out, greatly increasing and simplifying the software and compiler development effort. Having such an accurate cost model of the hardware was very important to drive hardware/software co-design decisions at an early stage of product development.

Finally, task descriptions and cycle times are formatted to conform to a prescribed schema and submitted as records to a database. The top half of figure \ref{phases_fig} provides a high-level view of this workflow. 

Once the database has enough submissions for an initial deployment of the cost model, the values are downloaded to a single location, and using the TensorFlow framework, the training phase of a neural network is initiated. This trained neural network can be versioned and released for consumption. VPUNN provides two different versions of the cost model that can be used by a VPU NN compiler. These two versions of VPUNN allow a compiler to choose between accuracy and speed depending on the preference of the host compiler and the application requirements. Both model versions expose the same Application Programming Interface (API) so the choice is completely transparent to the end-user of the library. Figure \ref{phases_fig} shows how the model can be deployed and used by the VPU compiler.

\subsection{Data preprocessing and feature engineering}

Initially, there are 26 raw input features which are then preprocessed and transformed into the final 71 features. The preprocessing step consists of categorically encoding the features that do not represent values, and logarithm the remaining numeric features for normalization. A more detailed analysis of the categorical encoding can be seen in Table \ref{categorical_table}. After these transformations have been done, the outlier data points are removed from the dataset before doing the train-validation split.

The output data is also transformed: The network is not trained to predict cycles values, which take values in the 100--10M range, but the utilization rate, which takes values in the 0--1 interval for dense workloads, a value that is much easier for the network to learn. The computed utilization rate can then be used to obtain the cycle value by understanding the cycles in a perfect utilization rate scenario.

\subsection{VPUNN model variants}
\subsubsection{Neural Network using Multi Layer Perceptron (MLP)}

\begin{figure}
  \centering
  \includegraphics[width=0.9\columnwidth, bb=-60 0 479 223]{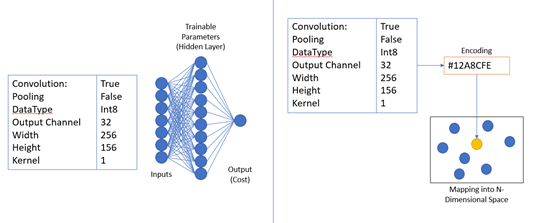}
  \caption{High level concept: (i)MLP method (ii) Embedding method}
  \label{Matrix Mul Diagram}
\end{figure}

For scenarios where a compiler may require a low-latency cost model, it can choose the VPUNN distribution based on an MLP network topology. This neural network has an input for each relevant aspect of a hardware task's description, has trained its parameters based on the database entries, and predicts cycle cost as an output. A simplified diagram can be seen in Figure \ref{Matrix Mul Diagram}(i). In addition to other performance-enabling features such as caching, this method can save a lot of time for a compiler - considering that the cost model may need to be queried tens of thousands of times for a single network compilation (due to a large number of tasks and configuration options as described in \ref{background_info}). Here, VPUNN takes advantage of the property of MLP network topologies to be converted to matrix-multiplication equivalents which can often take advantage of processor SIMD or vector instructions (\citet{conv_as_matmul}).

\subsubsection{Neural Network using Embeddings}

During training, the second style of the VPUNN model embeds costs from the database in an N-dimensional space, where each dimension corresponds with a property of the dataset (e.g., output channels). A simplified depiction of this insertion can be seen in Figure \ref{Matrix Mul Diagram}(ii). 

When an inference of the neural network occurs, it will either return an exact value of a previous recording or interpolate between the nearest neighboring values to return an approximation. This technique is inspired by the similarity searches used by \citet{Guo2020AcceleratingLI} and \citet{johnson}. This style of deployment provides higher confidence results due to its ability to return an exact match (where one exists), as opposed to the MLP  method which can only return predictions.

\begin{figure}
\centering
\subfloat{{\includegraphics[width=0.15\columnwidth,height=6cm, bb=0 0 174 764]{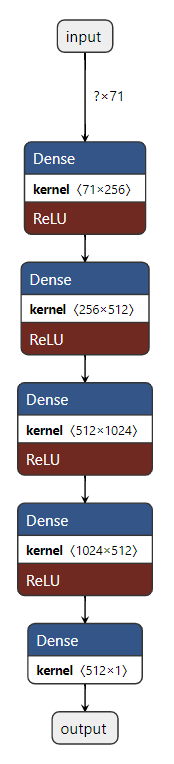} }}
\hspace{0.1\linewidth}
\subfloat {{\includegraphics[width=0.15\columnwidth, height=5cm, bb=0 0 174 572]{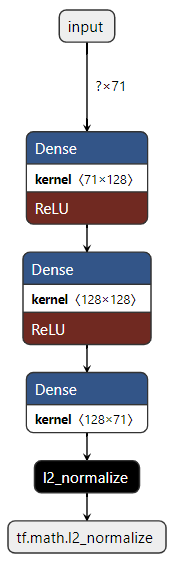} }}
\caption{Network Architectures: (i)MLP method (ii) Embedding method}
\label{architectures}
\end{figure}

\section{Results \& Discussion}

\subsection{Training VPUNN to approximate the hardware model}
\label{hyperparams}

\begin{figure}
  \centering
  \includegraphics[width=\linewidth, bb=-350 0 2163 833]{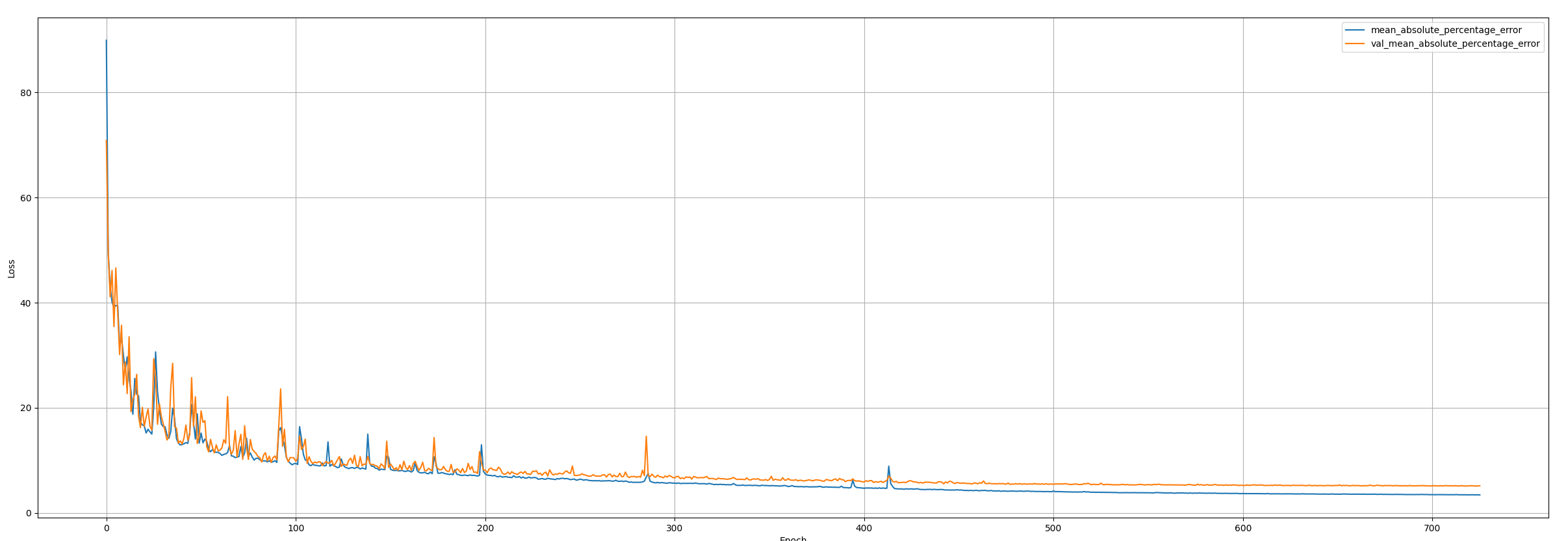}
  \caption{MAPE curve for matrix multiplication network}
\label{AC}
\end{figure}

The VPUNN model analyzed in this paper was trained on 31366 recordings of simulated workloads, using 80\% of the data for training and 20\% for validation.
The network was trained on an NVIDIA A100-SXM4 40GB GPU. On this system the training process takes 36 minutes - The network quickly learns features in the first few iterations and slowly refines them until we no longer see a significant benefit. The MAPE curve for the network can be seen in Figure \ref{AC}.

The Early Stopping callback is used, with a patience of 100 epochs and a minimum delta of 0.1. The network is trained with a batch size of 64, for 1000 epochs, however, it only requires approximately 700 epochs to fully train, when the training stops because of the callback. These hyperparameters were obtained through repeated experiments and proved themselves to offer the best validation results.

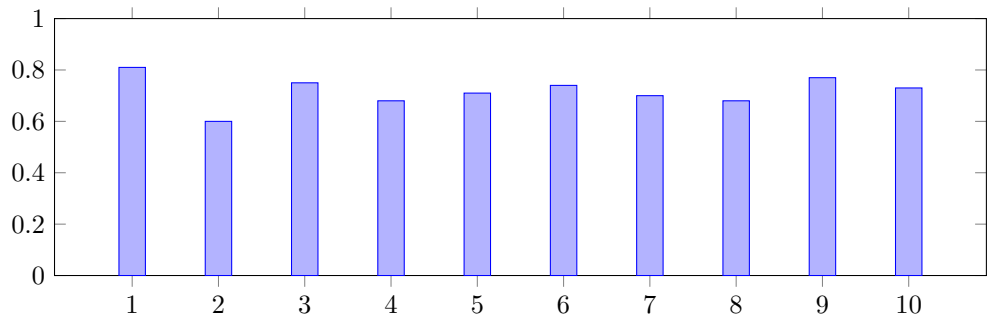
\begin{figure}[h]
\begin{tikzpicture}
    \begin{axis}[
            ybar,
            xtick=data,
            ymin=0,
            ymax=1,
	width=\textwidth,
	height=5cm,
        ]
        \addplot table[x=sample,y=median_ape]{\experimentsdata};
    \end{axis}
\end{tikzpicture}
  \caption{Median validation error bars (y) with respect to experiments with different seeds (x)}
  \label{experiments_table}
\end{figure}

At the end of the training, our model has a median validation Absolute Percentage Error (APE) of 0.72\%. This signals a high rate of confidence in the modeling approximation, and the resulted performance has high reproducibility throughout multiple experiments with different seeds as seen in Figure \ref{experiments_table}.

The Embedding version of VPUNN has been trained in the same fashion, however, there is a significant improvement in results. The Embedding model has a median validation APE of 0.31\%. As expected, this variant performs better.

\subsection{Effectiveness as a cost model}
\label{EF}

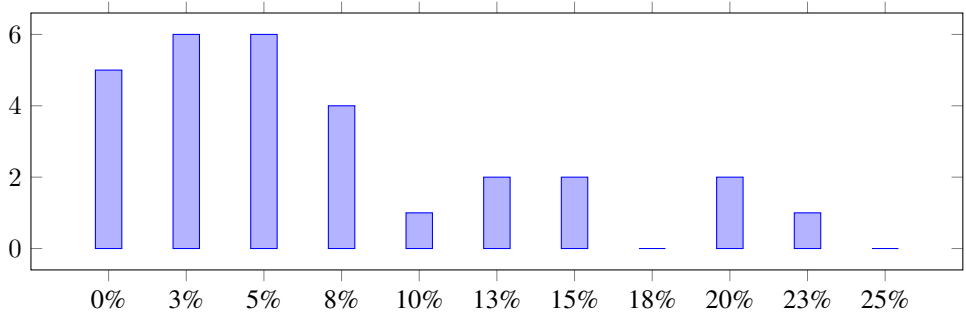
\begin{figure}
  \begin{tikzpicture}
    \begin{axis}[
        ybar,
        symbolic x coords={0\%,3\%,5\%,8\%,10\%,13\%,15\%,18\%,20\%,23\%,25\%},
        xtick=data,
        width=\textwidth,
        height=5cm,
      ]
      \addplot table[x=interval,y=amount]{\mydata};
    \end{axis}
  \end{tikzpicture}
  \caption{Observed improvements (x) in a quantity of neural networks(y)  }
  \label{vpux_table}
  
\end{figure}

The fact that the VPUNN model is comparable to the hardware modeling does not necessarily imply a guarantee of improved cost modeling and resultant task schedules. For instance, a poor hardware modeling may be an inferior cost model when compared to a mathematical formula that better correlates to the costs in the system. The only way to verify whether it is a better cost model is by direct comparison of the VPU's performance when using VPUNN in place of the existing, rule-based, methodology. Figure \ref{vpux_table} shows the improvements in VPU execution time for a range of networks (in Frames per Second (FPS)) due to the integration of the VPUNN cost model inside the compiler.

The findings are significant in that every neural network that was profiled either improved or did not degrade. This means that the cost model was capable of guiding the compiler to make better decisions than the previous rule-based approach.

\pgfplotstableread[row sep=\\,col sep=&]{
  interval & amount & carD & carR \\
  0\%     & 5  & 0.1  & 0.2  \\
  3\%     & 6 & 3.8  & 4.9  \\
  5\%    & 6 & 10.4 & 13.4 \\
  8\%   & 4 & 17.3 & 22.2 \\
  10\%   & 1  & 21.1 & 27.0 \\
  13\%      & 2  & 22.3 & 28.6 \\
  15\%      & 2  & 22.3 & 28.6 \\
  18\%      & 0  & 22.3 & 28.6 \\
  20\%      & 2  & 22.3 & 28.6 \\
  23\%      & 1  & 22.3 & 28.6 \\
  25\%      & 0  & 22.3 & 28.6 \\
}\mydata

\subsection{Execution Time}
\label{ET}

One might have been able to achieve similar results if they had used the complex hardware modeling directly as a cost model. However, these models are often very time-consuming and would be too slow for a standard user of the VPU compiler.

Table \ref{ms_table} shows the time to query both the complex hardware model and VPUNN for the different costs through a selection of networks. It can be seen that VPUNN takes far less time to return a result. Without this level of performance, using VPUNN as a cost model would have limited application, perhaps used as an optimization toggle to the compiler, but it is fast enough to be considered as a standard cost model for a compiler.

\begin{table}
  \caption{VPUNN execution times versus hardware model}
  \label{ms_table}
  \centering
  \begin{tabular}{lll}
    \toprule
    Network      & MLP VPUNN & Embeddings VPUNN \\
    \midrule
    ResNet-50    & 28.9x   & 21.1x              \\
    MobileNet v2 & 65.9x   & 31.2x              \\
    Yolo Tiny v2 & 47.8x   & 31.5x              \\
    \bottomrule
   \end{tabular}
\end{table}

Because of the widespread improvements were seen in Sections \ref{EF} and \ref{ET}, VPUNN has since been established as the primary cost model for all future VPU products.

\subsection{Limitations}
\label{limits}

There are two main limitations to the VPUNN approach to cost modeling. Firstly, VPUNN cannot be more accurate than the modeling of hardware it is approximating. The VPU target used in this paper was the third generation VPU in which the gathering of task-level profilings on hardware doesn't allow a zero observer effect. This made the profiling of individual workloads a difficult and error-prone task. An accurate modeling of the VPU runtime has been done to mitigate the effect of measurement on task profiling but the whole process is tricky at best to fine-tune.

To overcome these limitations, future VPU generations have a cycle-accurate hardware timer specifically tailored for individual task profiling. We include some preliminary findings in Appendix \ref{A_AB}, strongly indicating that using cycle costs from the actual hardware significantly reduces the impact of this limitation.

The second limitation is that gaps in training data could result in bad cost predictions by VPUNN. It is important, therefore, to have a training data set that covers as much of the network's design space in as much detail as possible. This large problem space needs a quite large training set to enable the network to properly generalize across unseen workloads.

\section{Conclusion and Future Plans / Use}
There is a clear improvement in the performance of the VPU when task schedules are compiled using VPUNN over the existing mathematical methods. VPUNN has been integrated into the OpenVINO VPU plugin because it is a very accurate model of complex hardware without the performance overhead included in a transaction level model.

A significant benefit of VPUNN is its enhanced ability to generalize to new and unseen networks (as discussed in Section \ref{EF}) and to efficiently scale to new workloads without much technical overhead. While an algorithmic model may require handcrafted analysis for every new type of workload, updating a neural network is much simpler and automatable once new workloads are collected in the database. It is possible to iteratively collect the data directly during pre-silicon validation and directly from end-users to allow the model to improve over time.

Though the topic of this paper has been focused on VPU, the concept is generic enough to apply VPUNN's successes to other targets provided the training is updated to appropriate task parameters. 

\subsection{Cost Model Tuning}
As the VPU moves through the different stages in its lifecycle - from design to delivery, the data used to train VPUNN can change too. Early-stage simulations or software modelings can be replaced with actual recordings on physical devices. This means that over time, releases of VPUNN for a target will be incrementally more accurate and representative. 

If data gathering is straightforward, software modeling may not even be needed. Provided with sufficient data gathering, VPUNN is very reusable between generations of a target, or even between different targets. To update the system, a new set of model parameters may need to be defined - whereas software modeling may need a lot of engineering time to understand the new properties of the system, assert the continued properties of continued features, debug mistakes, and so on. There is also a significant advantage in having a system less prone to the human element --- gathering extra training data is simpler than rewriting an entire software project!

\subsection{Dataset Exploration}
Outside of the improvements in cost model accuracy, having a singular repository of performance data is valuable for other uses for early-stage hardware developers. Exploration within the dataset can unveil unexpected performance characteristics, corner cases, or unmet expectations. Learnings from such investigations can be fed back into different parts of the organization to inform architectural and hardware design decisions, as well as firmware development choices. Collecting these datasets will enable future VPU generations to better understandings about performance and power at a much earlier stage of their lifecycles.

\subsection{Future Work: Neural Architecture Search}
The gathered dataset of tasks for VPUNN can be reused for other machine learning initiatives. One prominent example is using Neural Architecture Search (as described by \citet{ZophL16}) to vary the parameters of neural networks to contain operations and tensors that better suit the hardware whilst not compromising on accuracy. The work by \citet{NAS_TPU} has shown this to be a successful technique for optimizing the power utilization and performance throughput of a given network structure for Google's Tensor Processing Units (TPUs)--- a processor with similar trade-off considerations as the Intel VPUs. \citet{hamza_and_friends} have some established work in this area on a previous version of the VPU.

\subsection{Future Work: Optimizing for different costs}
For performance, the cycle count is an effective metric to train our neural networks for. However, particularly for Intel's range of VPU and other processors in the same market, power can be an important factor in competitive comparisons. Should this work be extended to gather the power cost for each task, a compiler could produce schedules that are optimized for power, rather than speed --- or a hybrid combination.

\paragraph{Summary} 
The invention of VPUNN significantly improves the ability of the VPU compiler to produce the most performant task schedule for a target VPU. It provides the benefit of a complex hardware modeling system at a fraction of the execution overhead. Due to these merits, VPUNN is now the standard costing system for the VPU compiler, already integrated into the Intel OpenVINO toolkit.

\section{Acknowledgements}

% To be left blank until submission

\bibliographystyle{plainnat}
\bibliography{VPUNN_NeurIPS}

% \begin{ack}
%Use unnumbered first level headings for the acknowledgments. All acknowledgments
%go at the end of the paper before the list of references. Moreover, you are required to declare
%funding (financial activities supporting the submitted work) and competing interests (related financial activities outside the submitted work).
%More information about this disclosure can be found at: \url{https://neurips.cc/Conferences/2022/PaperInformation/FundingDisclosure}.

%Do {\bf not} include this section in the anonymized submission, only in the final paper. You can use the \texttt{ack} environment provided in the style file to autmoatically hide this section in the anonymized submission.
%\end{ack}

%References follow the acknowledgments. Use unnumbered first-level heading for
%the references. Any choice of citation style is acceptable as long as you are
%consistent. It is permissible to reduce the font size to \verb+small+ (9 point)
%when listing the references.
%Note that the Reference section does not count towards the page limit.
%\medskip

%%%%%%%%%%%%%%%%%%%%%%%%%%%%%%%%%%%%%%%%%%%%%%%%%%%%%%%%%%%%

%\input{sections/VPUNN_06_Checklist.tex}
%%%%%%%%%%%%%%%%%%%%%%%%%%%%%%%%%%%%%%%%%%%%%%%%%%%%%%%%%%%%

% Not Needed for Preprint
%\clearpage
%Submit The Following Seperately as Supplemental Material:
%\clearpage
%\setcounter{page}{1}
%\setcounter{linenumber}{1}
%\appendix

\section{Appendix}

\subsection{Future Work: Improved Accuracy }
\label{A_AB}

Comparison of the Absolute Percentage Error between previous modelling tools and VPUNN when trained on hardware recordings --- versus the actual cycle count of hardware recordings. These were gathered on early versions of Generation 4 Intel VPUs, as described in Section \ref{limits}.
These results shows promise of improved accuracy of results if VPUNN were updated to use real hardware timings rather than simulated timings.

\begin{figure}[ht]
  \centering
  \includegraphics[width=0.7\textwidth, bb=-50 0 307 263]{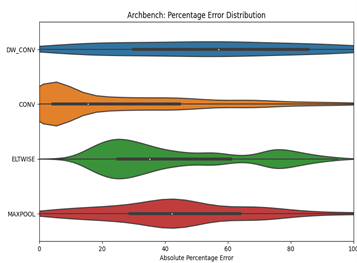}
  \caption{ArchBench absolute percentage error}
\label{MAPE_AB}
\end{figure}

\begin{figure}[ht]
  \centering
  \includegraphics[width=0.7\textwidth, bb=-50 0 307 263]{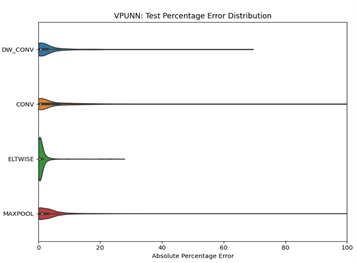}
  % \fbox{\rule[-.5cm]{0cm}{4cm} \rule[-.5cm]{4cm}{0cm}}
  \caption{VPUNN absolute percentage Error}
\label{MAPE_VPUNN}
\end{figure}

\subsection{Preprocessing}

Along with the fields presented here, there are some excluded fields set to zero, that are reserved for future versions of VPUs.

\FloatBarrier
\begin{table}[h]
  \caption{Existing values for categorical features}
  \label{categorical_table}
  \centering
  \begin{tabular}{ll}
    \toprule
    Feature name & Values \\
    \midrule
    input datatype   &  [uint8; int8; float16; bfloat16]          \\
    weights datatype   &  [uint8; int8; float16; bfloat16]         \\
    output datatype   &  [uint8; int8; float16; bfloat16]          \\
operation & \multirow{2}{10cm}{[Convolution; CM Convolution; DW Convolution; Elementwise; MaxPool; AveragePool]}\\ 
&  \\
    execution mode & [Vector; Matrix; Vector FP16; Cuboid 16x16; Cuboid 8x16; Cuboid 4x16] \\
    activation function & [None; RELU; LRELU; Add; Sub; Mult] \\
    \bottomrule
  \end{tabular}
\end{table}
\FloatBarrier

\subsection{Error Bars }

\FloatBarrier
\pgfplotstableread[row sep=\\,col sep=&]{
    sample & median_ape \\
    1     & 0.81  \\
    2     & 0.6   \\
    3     & 0.75  \\
    4     & 0.68  \\
    5     & 0.71  \\
    6     & 0.74  \\
    7     & 0.7   \\
    8     & 0.68  \\
    9     & 0.77  \\
    10    & 0.73  \\
    }\experimentsdata

\FloatBarrier

\subsection{Networks used for Assessment}

Details of the networks used for assessing VPUNN in comparison to the existing cost modelling within the VPU compiler. All networks were found to be either improved or unchanged in performance when VPUNN was used. 

\FloatBarrier
\begin{table}[h]
  \caption{Networks analysed for performance improvement/degradation}
  \label{networks_list}
  \centering
  \begin{tabular}{lll}
    \toprule
    Network      & Precision & Hyperlink \\
    \midrule

\detokenize{por_tf_FP16-INT8_yolo-v2-tiny-ava-0001} & Mixed FP16-INT8 & \href{https://docs.openvino.ai/latest/omz_models_model_yolo_v2_tiny_ava_0001.html}{OpenVINO Model Info} \\
\detokenize{por_tf_FP16-INT8_yolo_v4} & Mixed FP16-INT8 & \href{https://docs.openvino.ai/2021.2/omz_models_public_yolo_v4_tf_yolo_v4_tf.html}{OpenVINO Model Info} \\
\detokenize{por_tf_FP16-INT8_yolo_v4_tiny} & Mixed FP16-INT8 & \href{https://docs.openvino.ai/latest/omz_models_model_yolo_v4_tiny_tf.html}{OpenVINO Model Info}  \\

\detokenize{BDK_FP16-INT8_resnet-50-pytorch-sparse50} & Mixed FP16-INT8 & \multirow{2}{3cm}{Internal variant of  \href{https://docs.openvino.ai/latest/omz_models_model_resnet_50_pytorch.html}{OpenVino model}}\\ 
& &  \\

\detokenize{por_onnx_FP16-INT8_resnet-50-pytorch} & Mixed FP16-INT8 & \href{https://docs.openvino.ai/latest/omz_models_model_resnet_50_pytorch.html}{OpenVINO Model Info} \\    
\detokenize{por_tf_FP16-INT8_googlenet-v1} & Mixed FP16-INT8 & \href{https://docs.openvino.ai/2021.2/omz_models_public_googlenet_v1_tf_googlenet_v1_tf.html}{OpenVINO Model Info} \\
\detokenize{por_tf_FP16_googlenet-v1} & FP16 & \href{https://docs.openvino.ai/2021.2/omz_models_public_googlenet_v1_tf_googlenet_v1_tf.html}{OpenVINO Model Info} \\
\detokenize{por_tf_FP16-INT8_googlenet-v3} & Mixed FP16-INT8 & \href{https://docs.openvino.ai/latest/omz_models_model_googlenet_v3.html}{OpenVINO Model Info} \\  
\detokenize{por_tf_FP16_googlenet-v3} & FP16 & \href{https://docs.openvino.ai/latest/omz_models_model_googlenet_v3.html}{OpenVINO Model Info} \\                    
\detokenize{por_caffe2_FP16_squeezenet1.1} & FP16 & \href{https://docs.openvino.ai/latest/omz_models_model_squeezenet1_1.html}{OpenVINO Model Info} \\                      
\detokenize{por_caffe2_FP16-INT8_squeezenet1.1} & Mixed FP16-INT8 &  \href{https://docs.openvino.ai/latest/omz_models_model_squeezenet1_1.html}{OpenVINO Model Info} \\ 
\detokenize{scale_caffe_FP16_mobilenet-v1-1.0-224} & FP16 & \href{https://docs.openvino.ai/latest/omz_models_model_mobilenet_v1_1_0_224_tf.html}{OpenVINO Model Info} \\                        
\detokenize{scale_caffe_FP16-INT8_mobilenet-v1-1.0-224} & Mixed FP16-INT8 & \href{https://docs.openvino.ai/latest/omz_models_model_mobilenet_v1_1_0_224_tf.html}{OpenVINO Model Info} \\        
\detokenize{por_onnx_FP16-INT8_mobilenet-v2} & Mixed FP16-INT8 & \href{https://docs.openvino.ai/latest/omz_models_model_mobilenet_v2.html}{OpenVINO Model Info} \\
\detokenize{ea_caffe_FP16_age-gender-recognition-retail-0013_ww14} & FP16 & \href{https://docs.openvino.ai/latest/omz_models_model_age_gender_recognition_retail_0013.html}{OpenVINO Model Info} \\
\detokenize{scale_caffe_FP16-INT8_age-gender-recognition-retail-0013} & Mixed FP16-INT8 & \href{https://docs.openvino.ai/latest/omz_models_model_age_gender_recognition_retail_0013.html}{OpenVINO Model Info} \\
\detokenize{scale_onnx_FP16_resnet-18-pytorch} & FP16 & \href{https://docs.openvino.ai/latest/omz_models_model_resnet_18_pytorch.html}{OpenVINO Model Info} \\
\detokenize{scale_onnx_FP16-INT8_resnet-18-pytorch} & Mixed FP16-INT8 & \href{https://docs.openvino.ai/latest/omz_models_model_resnet_18_pytorch.html}{OpenVINO Model Info} \\
\detokenize{scale_caffe_FP16_mobilenet-v2} &  FP16 & \href{https://docs.openvino.ai/latest/omz_models_model_mobilenet_v2.html}{OpenVINO Model Info} \\
\detokenize{scale_caffe_FP16-INT8_mobilenet-v2} & Mixed FP16-INT8 & \href{https://docs.openvino.ai/latest/omz_models_model_mobilenet_v2.html}{OpenVINO Model Info} \\
\detokenize{FP16-INT8_InternalModel1} & Mixed FP16-INT8 & Internal Model \\ %MODEL_A
\detokenize{INT8_InternalModel2_V2} & INT8 & Internal Model \\          %MODEL_E
\detokenize{INT8_InternalMode2_V3} & INT8 & Internal Model\\           %MODEL_E
\detokenize{scale_tf_FP16_mobilenet-v3-large-1.0-224} & FP16 & \href{https://docs.openvino.ai/latest/omz_models_model_mobilenet_v3_large_1_0_224_tf.html}{OpenVINO Model Info} \\
\detokenize{scale_tf_FP16-INT8_mobilenet-v3-large-1.0-224} & Mixed FP16-INT8 & \href{https://docs.openvino.ai/latest/omz_models_model_mobilenet_v3_large_1_0_224_tf.html}{OpenVINO Model Info} \\
\detokenize{por_tf_FP16_mobilenet-v3-small-1.0-224} & FP16 & \href{https://docs.openvino.ai/latest/omz_models_model_mobilenet_v3_small_1_0_224_tf.html}{OpenVINO Model Info}  \\
\detokenize{por_tf_FP16-INT8_mobilenet-v3-small-1.0-224} & Mixed FP16-INT8 &  \href{https://docs.openvino.ai/latest/omz_models_model_mobilenet_v3_small_1_0_224_tf.html}{OpenVINO Model Info} \\
\detokenize{FP32_InternalModel3} & FP32 & Internal Model \\

    \bottomrule
  \end{tabular}
\end{table}
\FloatBarrier
\textbf{Note:} The networks referenced here may be using internal releases of OpenVINO or be non-published variants of the published network. 

\textbf{Note:} weblinks accessed 2022-04-29
%\end{tabular}
%\end{center}

% Uncomment this to have the bibliography added in and not sorted
% \nocite{*}

%Optionally include extra information (complete proofs, additional experiments and plots) in the appendix.
%This section will often be part of the supplemental material.

\end{document}